\documentclass{article}
\usepackage{spconf,amsmath,graphicx,hyperref}
\usepackage{cite}
\usepackage{makecell}

\usepackage{algorithm}
\usepackage{algorithmic}
\usepackage{amsmath}
\usepackage{multirow} 
\usepackage{booktabs} 
\usepackage{array}    
\usepackage{tcolorbox} 
\usepackage{caption} 
\tcbuselibrary{breakable}
\usepackage{enumitem}
\usepackage{xcolor}   
\usepackage{multirow}
\definecolor{ForestGreen}{HTML}{009B55}
\usepackage{array} 
\usepackage{tabularx}
\usepackage{seqsplit}

\title{Compound-QA: A Benchmark for Evaluating LLMs on Compound Questions}

\name{
    Yutao Hou$^{1}$, Yajing Luo$^{1}$, Zhiwen Ruan$^{3}$, Hongru Wang$^{2}$, Weifeng Ge$^{4}$, Yun Chen$^{1}$, Guanhua Chen$^{3*}$\thanks{\ \ * Corresponding Author.}
    }
\address{
    $^1$Shanghai University of Finance and Economics,
    $^2$University of Edinburgh \\
    $^3$Southern University of Science and Technology, $^4$Fudan University
}

\begin{document}
\ninept
\maketitle
\begin{abstract}
Large language models (LLMs) demonstrate remarkable performance across various tasks, prompting researchers to develop diverse evaluation benchmarks. However, most benchmarks typically measure the ability of LLMs to respond to individual questions, neglecting the complex interactions in real-world applications. We introduce Compound Question Synthesis (CQ-Syn) to build Compound-QA, a benchmark targeting questions composed of multiple interrelated sub-questions. This benchmark is derived from existing QA datasets, annotated with proprietary LLMs, and verified by humans for accuracy. It encompasses five categories: Factual-Statement, Cause-and-Effect, Hypothetical-Analysis, Comparison-and-Selection, and Evaluation-and-Suggestion. It evaluates the LLM capability in terms of three dimensions, including understanding, reasoning, and knowledge. Evaluating nine open-source LLMs on Compound-QA reveals that their performance on compound questions is notably lower than on non-compound questions. We further explore strategies to enhance LLMs’ handling of compound questions, and our results show that these methods substantially improve models’ comprehension and reasoning abilities. The dataset is publicly available at \url{https://github.com/sustech-nlp/Compound-QA}. 
\end{abstract}


\section{Introduction}

Large language models (LLMs) have achieved remarkable success in natural language processing (NLP), demonstrating exceptional performance across a wide range of tasks due to their advanced language understanding, reasoning, and generation capabilities~\cite{gpt-4,LLaMA31,team2025gemma,qwen3}. Existing benchmarks evaluate these models' abilities across various dimensions ~\cite{IFEVAL,alpaca_eval}, such as understanding~\cite{adversarial_qa,alpaca_eval}, reasoning~\cite{hotpotqa,bean2024lingoly}, and knowledge~\cite{recall,pubmedqa}. 
However, these benchmarks primarily evaluate responses to individual questions or instructions, overlooking the complexity of real-world interactions~\cite{he2024can}.

\begin{figure}[t]
    \centering
  \includegraphics[width=0.47\textwidth]{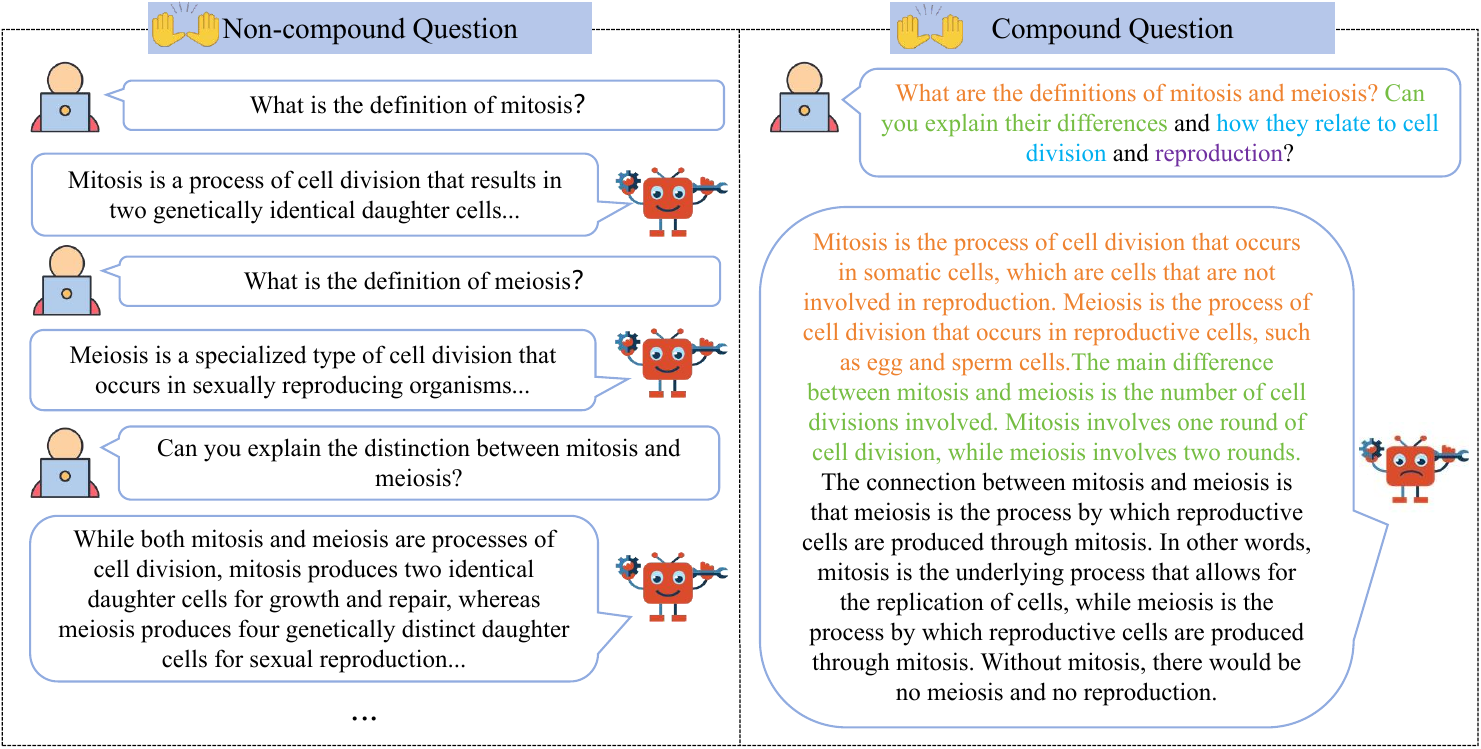}
 \vspace{-4pt}
  \caption{Examples of non-compound (left) and compound (right) questions: the former poses multiple questions across turns, while the latter combines them within a single turn.}
  \vspace{-15pt}
  \label{fig:compound-example}
\end{figure}

In real-world scenarios, users often ask a series of interrelated questions within a single query, expecting to obtain a comprehensive and precise response for each question, as illustrated in Figure~\ref{fig:compound-example}.
  
We refer to this as \textbf{Compound Questions}, which are queries that contain multiple sub-questions within one turn and may be correlated (Section~\ref{sec:cq_type}). 
This question format is common in human-AI interactions and agent-based scenarios, where tasks are decomposed into sub-instructions that require individual responses.
While humans can effectively address compound questions by answering each sub-question separately without omission or interference, LLMs face challenges such as identification of sub-questions and the elliptical phenomena in natural language \cite{van19elipsis}. Adjacent questions and answers can cause LLMs to focus on earlier context while overlooking unanswered sub-questions. Since LLMs are susceptible to irrelevant context, they may also be influenced by other sub-questions~\cite{wu2024easily}.

Recent studies explore how LLMs handle multiple-problem tasks~\cite{wang2024evaluating,liu2024longgenbench,chen2024sifo,wang2025evaluating,pan2025rest}. 
However, most studies focus on classification or fixed-answer tasks. Furthermore, the inter-question relationships they examine are generally simple, typically involving either concatenating questions~\cite{wang2024evaluating,liu2024longgenbench} or using sequential instructions where one answer influences the next~\cite{chen2024sifo,son2024multi}. In contrast, our work tackles both issues by targeting open-ended QA tasks and by abstracting the complex, real-world logical dependencies among sub-questions.

To this end, we introduce a data-synthesis framework, Compound Question Synthesis (CQ-Syn). This framework leverages LLM to generate and refine compound questions according to developed guidelines, followed by a thorough human review to ensure quality. Using CQ-Syn, we construct the Compound-QA to evaluate LLMs’ handling of compound questions. This benchmark consists of 1,500 samples spanning understanding, reasoning, and knowledge, categorized into five types: Factual-Statement, Cause-and-Effect, Hypothetical-Analysis, Comparison-and-Selection, and Evaluation-and-Suggestion. Experiments on nine open-source LLMs reveal that their effectiveness on compound questions is significantly lower than on single-question tasks, highlighting limitations in multi-step reasoning and contextual integration. However, supervised fine-tuning with instruction data augmented by compound questions substantially improves performance. We hope this work fosters further advancements in complex question answering.

\section{Compound-QA Benchmark} \label{sec:method}

\subsection{Types of Compound Questions} \label{sec:cq_type}

The compound question incorporates multiple sub-questions within a single query. It is common in human-AI interaction where users might propose several questions at one time. The ability to respond to compound questions is also important in agentic applications where several sub-instructions derived from the task decomposition and planning are expected to be followed. However, it is not trivial for LLMs as they might suffer from the problem of sub-question omission and a degradation in the quality of response (see examples in Figure~\ref{fig:compound-example}). 

The difficulty arises from several intrinsic characteristics of compound questions. The sub-questions may exhibit \textbf{hierarchical or logical dependencies}, requiring the model to not only understand each query individually but also grasp their interrelations. Furthermore, the context provided by one sub-question can create \textbf{contextual interference}, leading the model to produce irrelevant or imprecise responses for subsequent sub-questions. Finally, phenomena like anaphora or ellipsis can introduce \textbf{referential ambiguity}, making it challenging for the model to accurately resolve co-references.

To systematically analyze these challenges, we propose a typology that categorizes compound questions into five distinct types based on the dominant logical relationship between their sub-questions. Table~\ref{tab:compound_questions} provides illustrative examples for each category.

\begin{table}[ht]
    \centering 
    \captionsetup{font=small} 
    \renewcommand{\arraystretch}{0.8} 
    \setlength{\extrarowheight}{0.1pt} 

    \caption{Examples of different types of compound questions. The core ideas of each type are shown in \textit{italics}.}  
    \label{tab:compound_questions}
    \begin{tabular}{m{0.7cm} m{6.9cm}}
        \toprule    
        \textbf{Types} & \multicolumn{1}{c}{\textbf{Example}} \\    
        \midrule    
        \makecell[c]{FS} & What is your favorite \textit{sport}? Do you have any special skills or habits \textit{when playing sport}? \\    
        \midrule 
        \makecell[c]{CE} & \textit{Why} has online learning been able to spread rapidly in recent years? \textit{Based on these main reasons, what are the main impacts} of the popularization of online learning on society and individuals? \\    
        \midrule 
        \makecell[c]{HA} & \textit{If I am successful} in organizing this event, \textit{how} do I ensure that I maximize its \textit{impact}? \textit{If I don't}, \textit{how} do I deal with the potential negative \textit{consequences}?  \\    
        \midrule 
        \makecell[c]{CS} & \textit{Compared with} traditional classroom teaching, \textit{what are the advantages} of online learning modes in terms of enhancing learning efficiency and flexibility? \textit{What is the ideal way} to ensure the quality of teaching and student-teacher interaction? \\    
        \midrule 
        \makecell[c]{ES} & \textit{How do you evaluate the current operational status of} online education platforms? \textit{What are their advantages and disadvantages} in improving learning efficiency and meeting individual needs? Based on this, \textit{what measures} do you think these platforms should take in the future to promote continuous improvement and innovation? \\   
        \bottomrule    
    \end{tabular}%
    \vspace{-5pt}
\end{table}

\begin{itemize}[left=0pt, itemsep=-2pt]
    \item \textbf{Factual-Statement (FS):} Requires retrieving multiple, largely independent pieces of factual information. The sub-questions exhibit minimal correlation.
    \item \textbf{Cause-and-Effect (CE):} Tasks the model with first identifying the causes of an event or phenomenon and then explaining its resulting effects and impacts.
    \item \textbf{Hypothetical-Analysis (HA):} Presents a hypothetical scenario and requires an analysis of its potential outcomes under specified conditions or roles.
    \item \textbf{Comparison-and-Selection (CS):} Involves a comparative analysis of two or more items to identify similarities and differences, culminating in a selection based on specific criteria.
    \item \textbf{Evaluation-and-Suggestion (ES):} Tasks the model with first providing a critical evaluation of a subject by analyzing its strengths, weaknesses, and underlying mechanisms, and then proposing actionable suggestions for improvement.
\end{itemize}

\subsection{Data Collection} \label{sec:cq_sync}
\begin{figure}[t]
    \centering
  \includegraphics[width=0.45\textwidth]{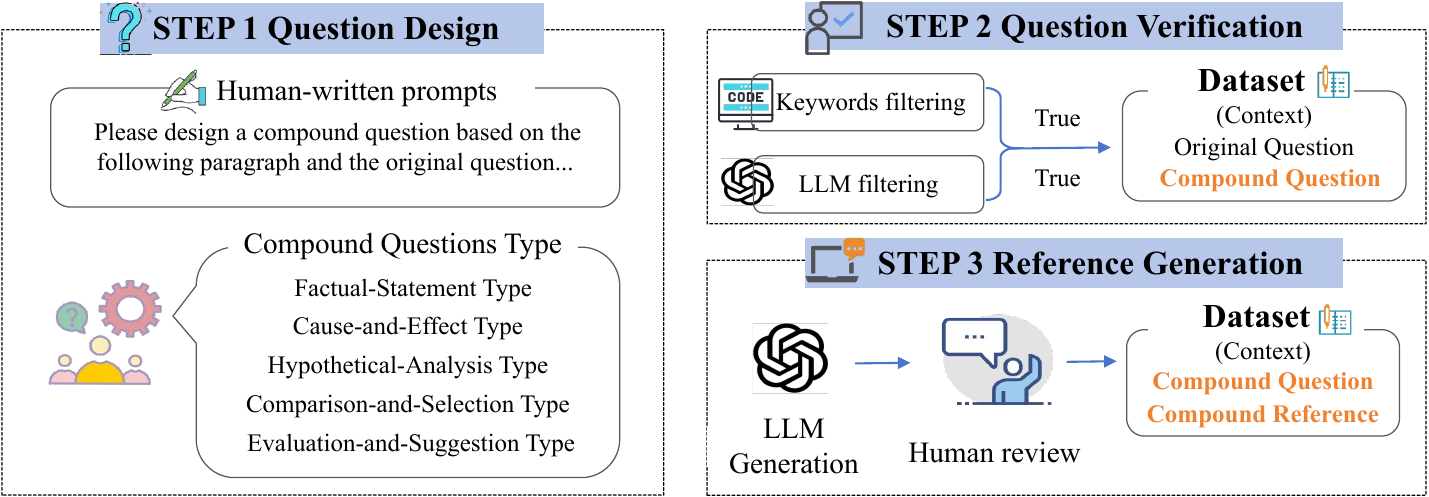}
    \vspace{-4pt}
  \caption{The overview of CQ-Syn Data Synthesis.}
  \vspace{-15pt}
  \label{fig:Dataset_construction}
\end{figure}

The Compound-QA is designed to evaluate an LLM's ability to handle compound questions. This task requires models to first comprehend and deconstruct the query, then address each sub-question sequentially and exhaustively.   
The benchmark comprises three subsets dedicated to understanding, reasoning, and knowledge, respectively. Each subset is constructed using existing related datasets (Adversarial QA~\cite{adversarial_qa}, AGI-Eval~\cite{agieval}, and PubMedQA~\cite{pubmedqa}) and includes compound questions of each type as detailed in Table~\ref{tab:compound_questions}. These questions are generated using our data synthesis framework, CQ-Syn, a three-phase process illustrated in Figure~\ref{fig:Dataset_construction}.

\begin{itemize}[left=0pt, itemsep=-2pt]
\item \textbf{Step 1: Question Design} We tailor unique prompts to address the characteristics of each type of compound question. The prompt covers task description, role description, and detailed data generation guidelines, along with manually curated examples to guide the process. We also include the corresponding context and the original question in the prompt, encouraging the LLMs to generate compound questions within a similar distribution of the original question based on the context. 
\item \textbf{Step 2: Question Verification} The generated compound questions are verified through both keyword-based and LLM-based filtering approaches. We apply keyword-based rules to filter out the generated questions that do not contain the pre-specified keywords. The rules are manually crafted for each type separately. 
\item \textbf{Step 3: Reference Generation} For each filtered compound question, we prompt the proprietary LLM to obtain the reference answers for each compound question. Each compound question and its corresponding reference answer are manually reviewed to ensure accuracy and quality. Human validation is conducted by three trained master's students using written guidelines, only samples unanimously judged complete and correct are retained.

\end{itemize}
\vspace{-10pt}

\subsection{Data Statistics}

Our Compound-QA dataset consists of three main subsets: understanding, reasoning, and knowledge. Each subset includes five types of compound questions: Factual-Statement, Cause-and-Effect, Hypothetical-Analysis, Comparison-and-Selection, and Evaluation-and-Suggestion. We generate 100 data points per type in each subset (5 types × 3 subsets), resulting in 1,500 data points in total. We use \texttt{gpt-4o} to generate the QA pairs.

\section{EXPERIMENTS}

\subsection{Experiment Setup}
We systematically evaluate nine open-source LLMs of different scales on the Compound-QA:
DeepSeek (\texttt{DeepSeek\allowbreak-llm\allowbreak-7b\allowbreak-chat})~\cite{DeepSeek-llm},
Mistral (\texttt{Mistral\allowbreak-7B\allowbreak-Instruct\allowbreak-v0.3})~\cite{Mistral},
LLaMA (\texttt{Meta\allowbreak-LLaMA\allowbreak-3.1\allowbreak-8B\allowbreak-Instruct})~\cite{LLaMA31}, Gemma-2 (\texttt{gemma\allowbreak-2\allowbreak-9b\allowbreak-it})~\cite{team2024gemma},
GLM-4 (\texttt{glm\allowbreak-4\allowbreak-9b\allowbreak-chat})~\cite{glm2024chatglm}, InternLM (\texttt{internlm2\allowbreak-5\allowbreak-7b\allowbreak-chat})~\cite{internlm2},
Qwen2.5 (\texttt{Qwen2.5\allowbreak-7B\allowbreak-Instruct})~\cite{qwen2.5}, 
Gemma-3 (\texttt{Gemma\allowbreak-3\allowbreak-27b\allowbreak-it})~\cite{team2025gemma}, and Qwen-3 (\texttt{Qwen3\allowbreak-32B})~\cite{qwen3}.

To quantify the model's ability in handling compound questions, we propose a multi-dimensional evaluation framework based on Comprehensiveness (explicit and complete addressing of all sub-questions), Correctness (factual and logical accuracy of each response component), and Diversity (variety in solution strategies across sub-questions). We use \texttt{gpt-4o-mini} as an automatic evaluator to compare model responses with reference answers and compute win rates as the percentage of instances in which a model is judged equal or superior to the reference. To mitigate position bias, we compute the scores with the order of the model response and reference answer swapped, then take the average as the final score. Overall performance is reported as the mean score across the three dimensions. Finally, we validate the automatic evaluations against human judgments and observe an agreement accuracy of 84\%.
\vspace{-7pt}
\subsection{Experiment Results and Analyses}

\subsubsection{How do different LLMs perform when answering compound questions?} 
\vspace{-7pt}

\begin{table}[h]
    \centering
    \captionsetup{font=small}
    \setlength{\extrarowheight}{0.1pt}
    \small
    \caption{The win rates of different LLMs on Compound-QA.}
    \setlength{\tabcolsep}{3.2pt} 
    \begin{tabular}{@{}l l r r r r r r@{}}
    \toprule    
         Models & Capability & Overall & FS & CE & HA & CS & ES \\
    \midrule
        \multirow{3}{*}{DeepSeek} & Understanding & 13.1 & 25.8 & 12.8	& 5.5 & 17.8 & 3.5 \\
        & Reasoning & 8.4 & 8.3 & 7.2 & 5.1 & 14.0 & 7.2  \\
        & Knowledge & 12.0 & 25.3 & 9.8 & 5.5 & 14.3 & 5.3 \\
    \midrule
        \multirow{3}{*}{Mistral} & Understanding & 22.4 & 40.0 & 18.0 & 15.2 & 29.8 & 8.8  \\
        & Reasoning & 13.6 & 10.8 & 11.7 & 11.6 & 22.5 & 11.5  \\
        & Knowledge & 14.6 & 29.3 & 11.2 & 9.3 & 16.0 & 7.2\\
    \midrule
        \multirow{3}{*}{LLaMA} & Understanding & 27.2 & 31.8 & 22.5 & 39.3 & 22.8 & 19.6  \\
        & Reasoning & 22.7 & 15.5 & 22.0 & 21.7 & 32.5 & 21.8 \\
        & Knowledge & 25.8 & 32.5	& 26.2	& 23.5	& 18.8	& 27.8\\
    \midrule
        \multirow{3}{*}{Gemma-2} & Understanding & 32.7 & 29.0 & 26.3 & 43.2 & 39.2 & 25.7 \\
        & Reasoning & 26.5 & 18.3 & 24.7 & 16.0 & 45.7 & 27.7 \\
        & Knowledge & 25.0 & 22.7 & 25.3 & 32.3 & 19.2 & 25.5\\
    \midrule
        \multirow{3}{*}{GLM-4} & Understanding &56.0 & 51.3 & 64.2 & 62.7 & 63.8 & 37.8 \\
        & Reasoning &41.7 & 26.7 & 47.8 & 29.6 & 60.3 & 44.2  \\
        & Knowledge & 46.7 & 44.8 & 54.5 & 47.2 & 48.0 & 38.8\\
    \midrule
        \multirow{3}{*}{Qwen2.5} & Understanding & 53.5 & 56.7 & 53.4 & 56.7 & 59.5 & 41.2\\
        & Reasoning & 48.8 & 43.8 & 52.5 & 41.5 & 56.2 & 50.2\\
        & Knowledge & 53.1 & 55.2 & 55.8 & 53.8 & 50.3 & 50.2\\ 
    \midrule
        \multirow{3}{*}{InternLM} & Understanding & 56.3 & 54.8 & 62.3 & 60.3 & 61.2 & 43.0 \\
        & Reasoning & 48.2 & 36.6 & 53.2 & 40.0 & 59.8 & 51.2 \\
        & Knowledge & 52.9 & 50.0 & 62.3 & 54.7 & 48.2 & 49.3 \\  
    \midrule
        \multirow{3}{*}{Gemma-3}
        & Knowledge & 55.5 & 46.8 & 55.3 & 63.3 & 46.7 & 65.3 \\
        & Reasoning & 59.4 & 53.2 & 50.0 & 56.7 & 67.0 & 70.0 \\
        & Understanding & 52.7 & 45.8 & 42.2 & 69.7 & 47.2 & 58.8 \\
    \midrule
        \multirow{3}{*}{Qwen3}
        & Knowledge & \textbf{90.5} & 86.8 & 93.3 & 87.7 & 93.2 & 91.3 \\
        & Reasoning & \textbf{76.2} & 49.7 & 84.7 & 69.6 & 89.0 & 88.2 \\
        & Understanding & \textbf{88.5} & 83.2 & 92.0 & 90.5 & 90.5 & 86.2 \\
    \bottomrule
    \end{tabular}
    \vspace{-8pt}
    \label{tab:quality_evaluation}
\end{table}

Table~\ref{tab:quality_evaluation} compares the win rates of nine open-source LLMs on five types of compound questions across the understanding, reasoning, and knowledge dimensions. Qwen3 obtains the strongest overall performance across all three dimensions, with a large margin. Gemma-3 shows competitive performance, especially in reasoning, but lags behind Qwen3 in knowledge and understanding. InternLM and GLM-4 remain competitive in understanding, while other models exhibit clear weaknesses on certain compound question types, underscoring their limited reasoning and synthesis abilities.

For the five types of compound questions, models generally perform best on Factual-Statement questions. In contrast, more complex question types, such as Evaluation-and-Suggestion, pose significant challenges for most models (e.g., InternLM 43.0, Qwen2.5 41.2); Qwen3, however, substantially narrows this gap, demonstrating markedly improved performance on these harder categories. This pattern indicates that increased model scale and stronger emergent reasoning abilities materially improve handling of compound questions.

\subsubsection{How do LLMs perform in answering compound versus non-compound questions?} 
\begin{figure}[h]
  \centering
  \includegraphics[width=0.9\columnwidth]{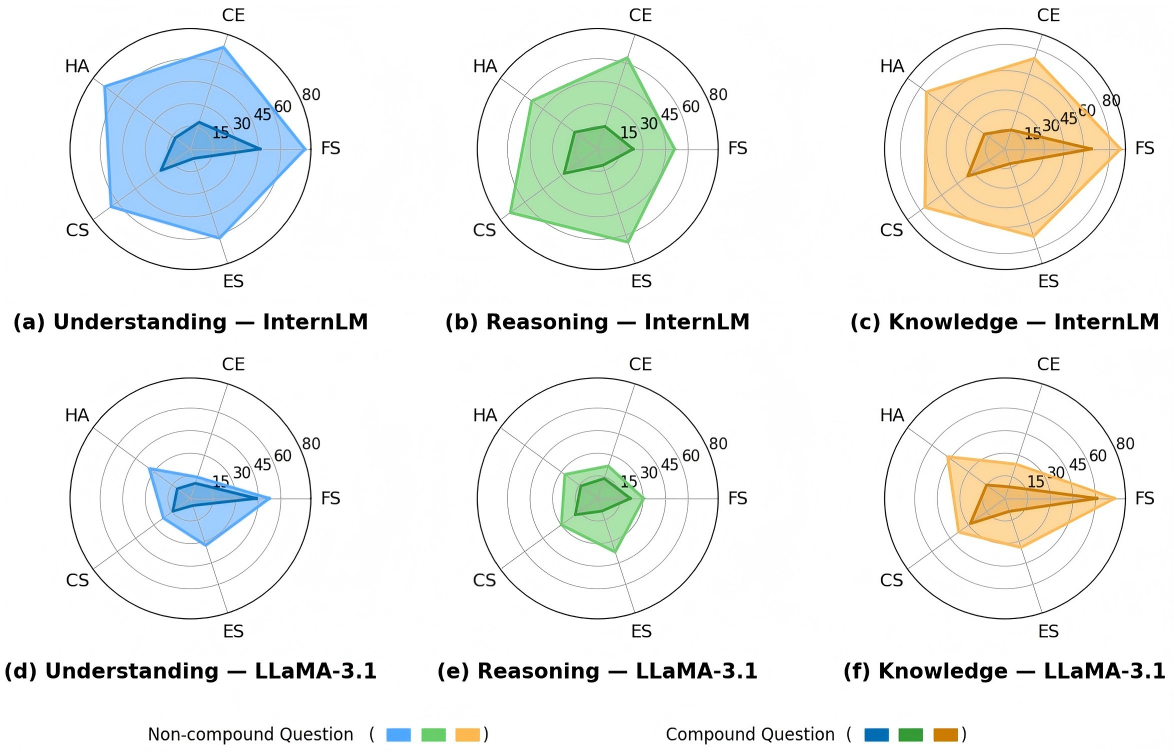}
  \caption{Performance comparison of LLaMA and InternLM when answering compound and non-compound questions.} 
  \vspace{-4pt}
  \label{fig:compound_non}
\end{figure}

To further understand model behavior in answering compound questions, we compare the performance of LLaMA and InternLM on both compound and non-compound questions. For non-compound questions, we use a multi-turn dialogue format, asking one question at a time. First, we decompose each compound question in our Compound-QA dataset into non-compound questions. Because some questions are challenging to decompose effectively, we perform a manual secondary check. Next, we use \texttt{gpt-4o} to generate reference answers for the decomposed questions. 
Finally, to ensure a fair comparison, we evaluate the models on the decomposed dataset by isolating their answers to each sub-question from compound responses and reporting win rates against the reference answers.

Figure~\ref{fig:compound_non} presents the win rates of LLaMA and InternLM on compound and non-compound datasets. The results clearly indicate that their performance drops substantially when handling compound questions. We find that the answers to compound questions are much shorter than those for non-compound questions, which could contribute to the decline in answer quality. Among the five compound question types, Factual-Statement questions are the least impacted by their compound structure, showing relatively strong performance in understanding and knowledge. This resilience likely stems from their simpler nature and weaker internal dependencies, as they often resemble a set of independent sub-questions. However, in the reasoning dimension, this pattern is less pronounced. Factual-Statement questions here often involve combinatorial logic reasoning, demanding a deep analysis of context to arrive at the correct response. This added complexity results in lower performance.

\subsubsection{How does LLM perform when answering sub-questions from different positions?} 

To evaluate LLM's performance in answering sub-questions from different positions, We reorder Factual-Statement questions within the understanding dimension and conduct experiments using LLaMA and InternLM. We choose the Factual-Statement questions because their sub-questions are relatively independent, and reordering them does not affect answers. Specifically, each data point contains three sub-questions, reordered so that each appears at the beginning, middle, and end of the sequence, labeled as 1XX, X1X, and XX1. We prompt the LLMs to answer these reordered compound questions.

Table~\ref{tab:different_positions} presents the win rates for answering sub-questions in different positions. The results show that sub-questions perform best when placed in the first and last positions, with the first position generally yielding the highest performance. Performance tends to decline in the middle position, suggesting that the model's ability to process information is stronger at the beginning and end of a sequence, but weaker in the middle. This is consistent with previous research findings that LLMs exhibit weaker performance when handling information located in the middle of long contexts~\cite{liu2024lost}.

\begin{table}[ht]
    \centering
    \renewcommand{\arraystretch}{0.8} 
    \captionsetup{font=small}
    \caption{Performance comparison of LLaMA and InternLM on sub-questions at different positions in compound questions.}
    \begin{tabularx}{\linewidth}{l@{\hspace{1pt}} l *{3}{>{\centering\arraybackslash}X} }
    \toprule
     Model & Sub-question & 1XX & X1X & XX1 \\
    \midrule
    \multirow{3}{*}{LLaMA} 
     & Sub-question1 & \textbf{40.7} & 38.8 & 37.7   \\
     & Sub-question2 & 38.7 & 40.8 & \textbf{41.8}   \\
     & Sub-question3 & \textbf{44.2} & 40.8 & 41.7 \\
    \midrule
    \multirow{3}{*}{InternLM} 
     & Sub-question1 & 45.0 & 42.5 & \textbf{47.5}  \\
     & Sub-question2 & \textbf{51.3} & 44.5 & 49.8  \\
     & Sub-question3 & \textbf{54.7} & 48.5 & 53.8  \\
    \bottomrule
    \end{tabularx}
    \vspace{-15pt}
    \label{tab:different_positions}
\end{table}

\subsubsection{How to improve LLM performance on Compound-QA ?} 


\begin{figure}[b]
  \centering
  \includegraphics[width=\columnwidth]{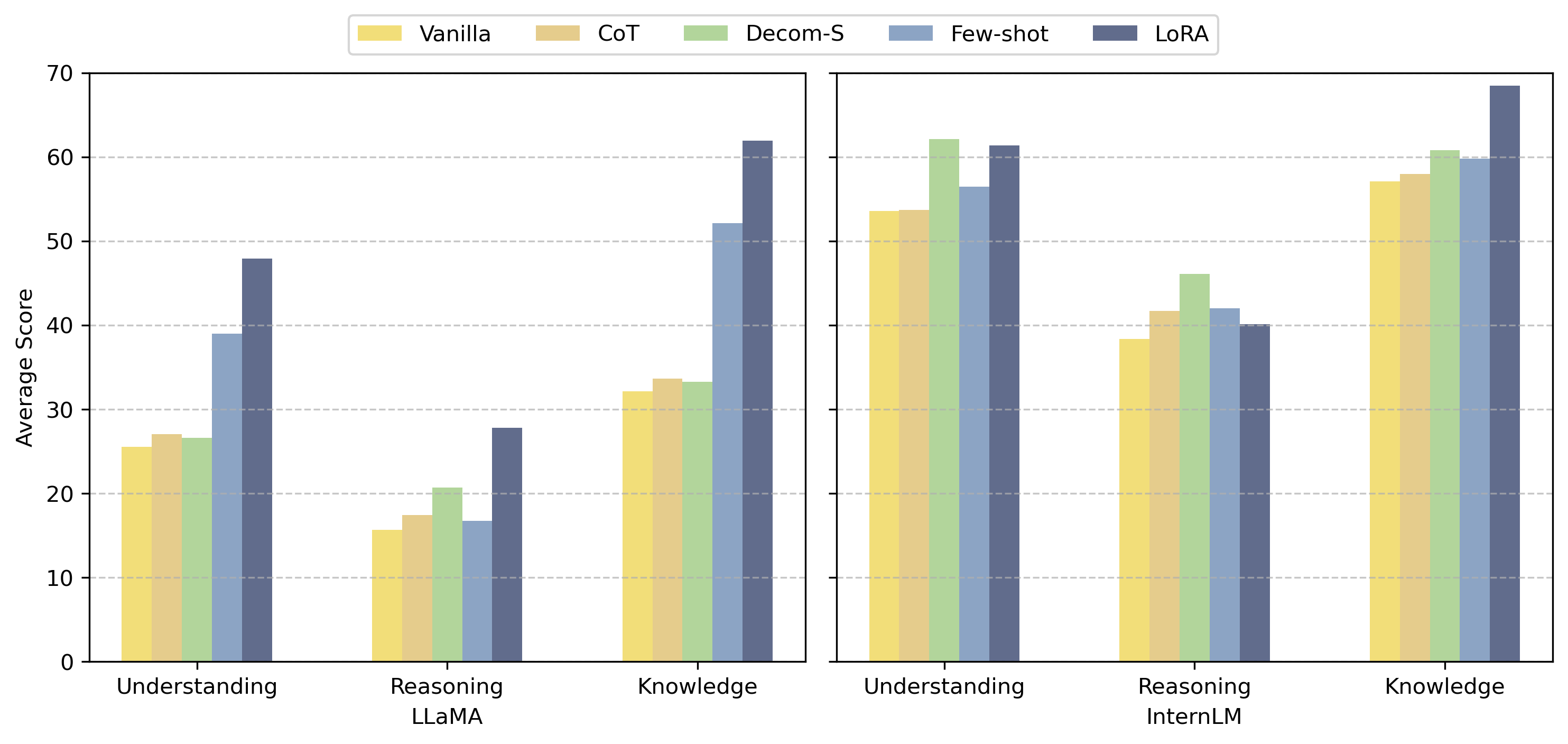}
    \vspace{-8pt}
  \caption{Comparative Performance of Different Improvement Methods on Compound-QA.} 
  \vspace{-5pt}
  \label{fig:lora_ft2}
\end{figure}

In this experiment, we investigate different approaches to enhance the model's ability to answer compound questions. We test the LLaMA and InternLM model using four methods: Chain-of-Thought (CoT), Decomposition strategy (Decom-S), Few-shot, and LoRA fine-tuning~\cite{hu2021lora}. CoT promotes step-by-step reasoning, whereas Decom-S explicitly instructs the model to break down a compound question into multiple sub-questions, address each sequentially, and synthesize the individual responses into a complete answer. To ensure a balanced evaluation across question types, the dataset is divided for each category and dimension (understanding, reasoning, and knowledge) in a 7:3 ratio. The 70\% portion of each category and dimension is combined to form the training set, while the remaining 30\% is used as the test set. CoT and Decom-S are evaluated in a zero-shot setting using the same test set. Figure~\ref{fig:lora_ft2} compares the model’s original and improved performance. LoRA fine-tuning achieves the highest scores on most dimensions, demonstrating its effectiveness for compound question answering.

Additionally, we further evaluate the fine-tuned model’s general capabilities on MMLU~\cite{mmlu}, GSM8K~\cite{cobbe2021gsm8k}, and TruthfulQA~\cite{truthfulqa} benchmarks using accuracy as the evaluation metric. As shown in Table~\ref{tab:generic}, the fine-tuned model maintains strong performance across these benchmarks, with no significant degradation in its ability to handle diverse tasks. This demonstrates that fine-tuning on Compound-QA preserves the model’s generalization ability while enhancing its performance on question answering.

\subsubsection{Error Analysis}

By analyzing responses of open-source models on the Compound-QA, we identify three typical errors: (1) Omission of sub-questions: the model fails to address all sub-questions and leaves part of the query unanswered; (2) Confusion between related sub-questions: the model conflates two or more closely related sub-questions and provides a blended or incomplete response (a behavior consistent with prior findings~\cite{chen2024sifo}); and (3) Off-topic responses: the model produces content that does not follow the logical reasoning steps required by the question and fails to reach a valid final answer. These observations highlight significant deficiencies in handling complex, multi-step reasoning tasks and offer valuable insights for future model refinements.

\section{Conclusion}

\begin{table}[t]
    \centering
    \renewcommand{\arraystretch}{0.8}
    \captionsetup{font=small} 
    \caption{Performance of LLaMA and InternLM models on generic tasks. Bold indicates the best score in each column.}
    \begin{tabularx}{\linewidth}{l *{3}{>{\centering\arraybackslash}X}}
    \toprule
     Models  & MMLU & GSM8K & TruthfulQA \\
    \midrule
     LLaMA & \textbf{67.78} & 83.00 & 47.61 \\
     LLaMA-LoRA & 67.73 & \textbf{84.50} & \textbf{52.39} \\
    \midrule
     InternLM & 70.59 & 74.00 & \textbf{54.95} \\
     InternLM-LoRA & \textbf{71.18} & \textbf{81.50} & 54.83 \\
    \bottomrule
    \end{tabularx}
    \vspace{-7pt}
    \label{tab:generic}
\end{table}

Compound questions present multiple sub-questions in a single query, which imposes challenges for LLMs to provide correct and appropriate responses to each sub-question in the Human-LLM interactive scenario.
We introduce Compound-QA, a benchmark designed to evaluate the ability of LLMs on compound questions. 
This benchmark categorizes compound questions into five different types, with each type covering the scenarios of understanding, reasoning, and knowledge. 
The dataset is created using a Human-LLM collaborative framework, which includes a data synthesis process for generating and verifying compound questions.
Our experiment reveals that LLMs require further improvement in effectively handling compound questions. We hope our benchmark will contribute to enhancing this capability. 
Additionally, we leave the exploration of evaluating compound questions in multimodal applications as future work.

\section*{Ethical Standards}

This study conducts numerical simulations. The dataset was manually reviewed to verify accuracy. No ethical approval was required.

\bibliographystyle{IEEEbib}
\bibliography{refs}

\end{document}